\documentclass[letterpaper, 10 pt, journal, twoside]{IEEEtran}
\IEEEoverridecommandlockouts 

\usepackage{amsmath} 
\usepackage{amssymb}
\usepackage[mathscr]{euscript}
\usepackage{graphicx}
\usepackage{xcolor}
\usepackage{algorithm}
\usepackage[noend]{algpseudocode}
\usepackage{tabularx}
\usepackage{multirow}
\usepackage{hhline}
\usepackage{geometry}
\usepackage{subcaption}
\usepackage{gensymb}
\usepackage[scientific-notation=true]{siunitx}

\usepackage{tikz}
\usepackage{textcomp}
\usepackage{hyperref}
\usepackage{lipsum}

\usepackage{dblfloatfix}

\usepackage[capitalise,nameinlink]{cleveref}
  \crefname{section}{Sect.}{Sect.}
  \Crefname{section}{Section}{Sections}
  \crefname{figure}{Fig.}{Fig.}
  \Crefname{figure}{Figure}{Figures}
  \crefname{table}{Tabl.}{Tabl.}
  \Crefname{table}{Table}{Tables}
\title{Accelerating Robot Learning of Contact-Rich Manipulations: A Curriculum Learning Study}
\begin{document}
\makeatletter
\newcommand{\linebreakand}{%
  \end{@IEEEauthorhalign}
  \hfill\mbox{}\par
  \mbox{}\hfill\begin{@IEEEauthorhalign}
}
\makeatother

\author{
 Cristian C. Beltran-Hernandez$^{{1}{*}}$ \quad
 Damien Petit$^{1}$ \quad
 Ixchel G. Ramirez-Alpizar$^{2,1}$ \quad 
 Kensuke Harada$^{1,2}$ %
\thanks{$^{1}$ Department of Systems Innovation, Graduate School of Engineering Science, Osaka University, Japan.}
\thanks{$^{2}$ Automation Research Team, Industrial CPS Research Center, National Institute of Advanced Industrial Science and Technology (AIST), Japan.}
\thanks{$^{*}$ Corresponding author e-mail: \newline \hspace*{10mm} {\tt cristian\_beltran[at]ieee.org}}
}
\maketitle


\begin{abstract}
The Reinforcement Learning (RL) paradigm has been an essential tool for automating robotic tasks. Despite the advances in RL, it is still not widely adopted in the industry due to the need for an expensive large amount of robot interaction with its environment.
Curriculum Learning (CL) has been proposed to expedite learning. However, most research works have been only evaluated in simulated environments, from video games to robotic toy tasks.
This paper presents a study for accelerating robot learning of contact-rich manipulation tasks based on Curriculum Learning combined with Domain Randomization (DR). We tackle complex industrial assembly tasks with position-controlled robots, such as insertion tasks. We compare different curricula designs and sampling approaches for DR. Based on this study, we propose a method that significantly outperforms previous work, which uses DR only (No CL is used), with less than a fifth of the training time (samples). Results also show that even when training only in simulation with toy tasks, our method can learn policies that can be transferred to the real-world robot. The learned policies achieved success rates of up to 86\% on real-world complex industrial insertion tasks (with tolerances of $\pm 0.01~mm$) not seen during the training.
\end{abstract}


\section{Introduction}

    Reinforcement Learning (RL) has been proven to be successful at learning complex behaviors to solve a variety of robotic contact-rich tasks \cite{sutton2018reinforcement,kober2013reinforcement,wang2020deep}.
    However, RL solutions are still not widely adopted in real-world industrial tasks. One reason is that RL still requires an expensive and large amount of robot interaction with its environment to learn a successful policy. The more complex the target task is, the more interaction (samples) is required.
    
    To tackle this problem, domain transfer methods such as Domain Randomization (DR) and Curriculum Learning (CL) have been introduced. 
    The concept of CL, where the learning process can be made more efficient by following a curriculum that defines an order in which tasks should be learned, has been introduced in previous works \cite{bengio2009curriculum, krueger2009flexible}. Additionally, DR of visual and physical properties of a task has been shown to improve the performance of tasks in novel domains \cite{tobin2017domain, andrychowicz2020learning}. However, most of these results have been only validated in simulated environments, from video games to robotic toy tasks, or in real-world toy environments. In this work, we tackle the problem of improving sample efficiency and performance when learning real-world complex industrial assembly tasks with rigid position-controlled robots. To this end, we seek to answer the question: Does the order in which the different environments (or tasks) are presented to the agent (through DR) affect the training sample efficiency and performance of the learned policy?
    
    We hypothesize that on top of DR, guiding a RL agent's training with a curriculum (presenting tasks in increasing order of difficulty) towards the desired behavior can increase sample efficiency. The reasoning is that the curriculum helps reduce the overall exploration needed to achieve the desired goal while DR enhances domain transferability.
    
    This paper presents a study of the combination of CL with DR. More specifically, we compare different curricula designs and different approaches at sampling values for DR. As a result, we propose a novel method that significantly outperforms our previous work \cite{beltran2020variable}. In \cite{beltran2020variable}, only DR is used to improve sim2real transferability without CL. Experimental results in simulation and real-world environments show that our novel method can be trained with only a fifth of the training samples required by our previous method and still successfully learn to solve the target insertion tasks. Furthermore, the learned policies transferred to the real world achieved high success rates (up to 86\%) on industrial level insertion tasks, with tolerances of $\pm 0.01~mm$, not seen during the training.
    
    This work's contributions are as follows:
    
    \begin{itemize}
        \item A study of the application of Curriculum Learning to a learning framework for rigid robots solving contact-rich manipulation tasks.
        \item A novel learning framework combining curriculum learning with domain randomization to accelerate learning and domain transfer.
        \item An improved reward function to guide the learning of force sensor-based contact-rich manipulation tasks. The reward perceived by the agent is dynamically discounted by the curriculum's level of difficulty. For our target domain, the idea is to encourage the agent to learn to solve the hardest tasks as discussed in \Cref{subsubsec:dynamic-reward}.
        \item An empirical study of the different methods considered in this paper was conducted. Novel tasks not seen during training were used to validate the performance of each method, both in simulation and in the real world, including complex industrial insertion tasks.
        Additionally, we study the impact of different components of our proposed method in the Appendix.
    \end{itemize}
    Additionally, this work's source-code\footnote{At https://github.com/cambel/robot-learning-cl-dr} will soon be open to the research community.
    
    The rest of this paper is organized as follows, related work is discussed in \Cref{sec:related-work}. The case study for this work and the proposed method are explained in \Cref{sec:materials-methods}. Experimental results and comparisons with alternative methods and our previous work are shown in \Cref{sec:experiments-results}. Ablation studies are described in the Appendix.

\section{Related Work} \label{sec:related-work}
    The related work to the topic of this paper, accelerating robot learning of contact-rich manipulation tasks, is introduced and discussed in this section.

\subsection{Reinforcement Learning for contact-rich manipulation tasks}
    Plenty of methods have been proposed to accelerate automation of robotic assembly tasks, such as peg-in-hole tasks. From search strategies to align the peg with the hole \cite{park2013intuitive,chhatpar2001search}, to learning-based methods \cite{luo2021learning, yasutomi2021peg}. Some researchers have explicitly focused on the domain transfer of assembly tasks from simulation to real-world environments. In \cite{schoettler2020meta}, a meta-RL technique is applied to transfer experiences and generalize better to the real world. In \cite{kaspar2020sim2real}, system identification of the real robot (KUKA LBR iiwa) with its simulated counterpart is performed to improve sim2real transferability. While RL-based policies have been proposed and proven to have the potential to solve assembly tasks in the real world, there is still a lack of adoption of such methods in real industrial assembly tasks. One reason for this gap between research and industry is the sample efficiency of such learning methods; a large amount of interaction of the agent with its environment is still necessary to learn robust policies. We aim to contribute to this area by proposing a more sample-efficient approach based on CL and DR, i.e., less time is required to train a successful policy without decreasing its transferability.

\subsection{Domain Randomization}
In the context of machine learning, DR has been proposed as a technique to improve domain transfer, such as going from one task to a harder one or moving from a simulated environment to a real-world environment, in particular for training vision-based models \cite{tobin2017domain} or sim2real models \cite{andrychowicz2020learning}. In \cite{mehta2020adr} an empirical study is presented to examine the effects of DR on agent generalization. Their results show that DR may lead to suboptimal, high-variance policies, which \cite{mehta2020adr} attributes to the uniform sampling of environment parameters. Following those results, this study proposed the use of DR based on CL and empirically studied the effect of the DR's sampling strategy.

\subsection{Curriculum Learning}
The concept of curriculum learning in the context of machine learning was first proposed by Bengio et al. \cite{bengio2009curriculum}. As mentioned before, CL can be understood as learning from easier to harder tasks, i.e., the order in which information is presented affects the policy's ability to learn. A comprehensive survey on Curriculum Learning applied to Reinforcement Learning has been presented in \cite{narvekar2020curriculum}. 
Most CL approaches have been validated mainly on simulated environments, such as toy examples (e.g., grid worlds, cart-pole, and low-dimensional environments), video games, and simulated robotic environments. Few research works have focused on real-world robotic environments, such as in \cite{asada1996purposive} where a robot is trained to shoot a ball into a goal, \cite{baranes2013active, luo2020accelerating} where reaching tasks with a robot arm are tackled, and \cite{riedmiller2018learning} which focused on two tasks moving a cube to a target pose and cube stacking. Most recently, \cite{leyendecker2021deep} presented a CL method focused on a specific automotive production task, trained on simulation, and transferred to its real-world equivalent. Similarly, \cite{yasutomi2022curriculum} proposes a method to enable a robot to conduct anchor-bolt insertion, a peg-in-hole task for holes in concrete.
On the other hand, our study focuses on tackling various real-world complex industrial assembly insertion tasks, trained only on toy peg-in-hole simulated environments. Furthermore, we make an exhaustive study on the performance of several approaches to combine DR and CL. As a result, we propose a method to accelerate learning and domain transfer to real-world environments by an adaptive curriculum that affects how DR and the reward signal are considered during training. 
\begin{figure*}[t]
    \centering
    \includegraphics[width=\textwidth]{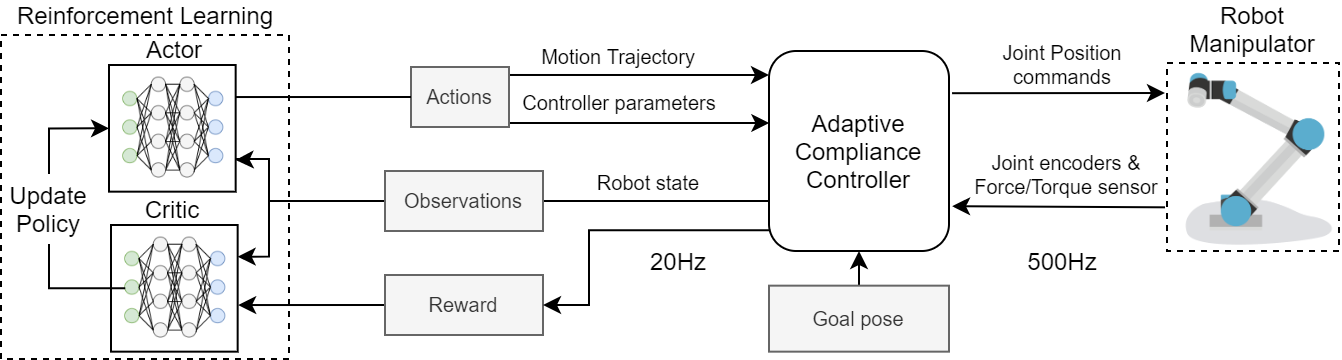}
    \caption{Overview of the system used for this study. The input is the goal pose, optionally the desired contact force can be defined, otherwise is considered as $0$~N. }
    \label{fig:system-overview}
\end{figure*}

\color{black}
\section{Materials and Methods}\label{sec:materials-methods}

\subsection{Problem Statement}\label{sec:problem-statement}
 
    In the present study, we consider the peg-in-hole assembly task that requires the mating of two components. One of the components is grasped and manipulated by the robot manipulator, while the second component has a fixed position via fixtures to a support surface. The proposed method is designed for position-controlled robot manipulators with access to force/torque information at the robot's end-effector (e.g., F/T sensor at the robot wrist), especially those robots where low-level torque control is not available. Thus, sensor-based force control is necessary to realize contact-rich manipulation tasks for such a type of robot.
    
\subsection{System Overview}\label{subsec:system-overview}

    Our propose method aims to improve the sample efficiency of the training phase.   \Cref{fig:system-overview} shows the overall system architecture which is based on our previous work \cite{beltran2020variable}. There are two control loops. The inner loop has an adaptive compliance controller; we choose to use a parallel position-force controller that was proven to work well for this kind of contact-rich manipulation tasks \cite{beltranhern2020learning}. The inner loop runs at a control frequency of 500 Hz, which is the maximum available in the Universal Robots e-series robotic arms\footnote{Robot details at https://www.universal-robots.com/e-series/}. The outer loop runs at a lower control frequency to account for the computation time required by the learning algorithm. Our system considers control of the 6 degrees of freedom of the Cartesian space at the robot's end-effector (position and orientation). To this previous learning control framework, we added the PID gains scheduling approach discussed in \Cref{subsubsec:pid-gains-scheduling}. Additionally, a new dense reward function is proposed and described in \Cref{subsubsec:cost-function}. Similarly, a DR method based on CL is implemented on top of this learning control framework, see \Cref{subsubsec:domain-randomization} and \Cref{subsubsec:curriculum-learning}.

\subsection{Reinforcement Learning}\label{subsec:rl-algo}
    The reinforcement learning framework followed in this work is modelled as an episodic Markov Decision Process (MDP), $M$, that has finite time steps with a limit of $T$ steps per episode. For a given task, the MDP consists of a state $\textbf{s} \in \mathscr{S}$, action space $\textbf{a} \in A$, state transition function $p(\textbf{s}(t+1) \,|\, \textbf{s}(t), \textbf{a}(t))$, which is the probability of transitioning to state $s'$ given that the action $a$ is taken while being in state $s$, and a reward function $r(s, a)$, which provides an immediate numerical reward for being in state $s$ and taking the action $a$. The goal of RL is to find a policy $\pi^*$ that maximizes the expected sum of discounted future rewards given by $R(t)=\sum_i^{\textbf{T}} \gamma r(s(t),a(t))$, where $\gamma$ is the discount factor \cite{sutton2018reinforcement}.
     
    We used a soft actor-critic (SAC) approach to learning the policy. SAC \cite{Haarnoja2018SoftAO} is an off-policy actor-critic deep RL algorithm based on maximal entropy. SAC aims to maximize the expected reward while also optimizing maximal entropy. The SAC agent optimizes a maximal-entropy objective, encouraging exploration according to a temperature parameter $\alpha$. The core idea of this method is to succeed at the task while acting as randomly as possible. Since SAC is an off-policy algorithm, it uses a replay buffer to reuse information from recent rollouts for sample-efficient training. Additionally, we used the distributed prioritized experience replay approach for further improvement \cite{horgan2018distributed}. Our implementation of the SAC algorithm was based on the TF2RL repository\footnote{TF2RL: Deep-reinforcement-learning library using TensorFlow 2.0. https://github.com/keiohta/tf2rl}.
    
    The agent's policy maps a multi-modal state, the robot's proprioception and the Force-Torque sensor data, to the robot's actions, detailed in \ref{subsec:parallel-control}, using a Time Convolutional Network (TCN); this policy representation and its improved performance over a simple neural network-policy were introduced in our previous work \cite{beltran2020variable}.
  
  \subsubsection{Reward function}\label{subsubsec:cost-function}
    On one hand, in our previous work \cite{beltran2020variable}, the reward function is defined only in terms of the contact force and distance between the current position of the robot's end-effector and the target position. The reason is to encourage the agent to get closer to the target position; the faster, the better, while discouraging any contact force. On the other hand, in this work, we propose the inclusion of the velocity of the robot's end-effector in the reward signal. While it is ideal that the agent achieves the task as fast as possible, high speeds are not desirable when the robot is close to the environment or in contact with it, as it can generate large contact forces. Thus, the proposed reward function has the following shape:
    
    \begin{equation}
    \begin{aligned}
        {r}(\textbf{s},\textbf{a}) = w_{1}r_{xv} + w_{2}r_F + w_3\rho
    \end{aligned}
    \label{eq:reward-function}
    \end{equation}
    where $r_{xv}$ is the component of the reward associated with the position and velocity of the robot's end-effector. $r_{xv}$ aims to encourage the agent to get closer and keep closer to the target position. Besides, the agent is encourage to move faster, if it is far from the target pose, or slower when it is close to the target pose. $r_{xv}$ is defined as:
    \begin{equation}
    \begin{aligned}
        r_{xv} = (1 - \tanh(5|x|)(1-|\dot{x}|) + (|\dot{x}|/2)^2
    \end{aligned}
    \label{eq:reward-position-velocity}
    \end{equation}
    Where $x$ is the distance between the robot's end-effector and the target position, and $\dot{x}$ is its velocity. A visualization of this reward component is shown in \Cref{fig:reward_xv}.
    The component of the reward \cref{eq:reward-function} associated with the contact force is defined as:
    \begin{equation}
    \begin{aligned}
        r_F = -1/(1+e^{-15|F_g-F_{ext}|+5})
    \end{aligned}
    \label{eq:reward-force}
    \end{equation}
    where $F_g$ is the desired insertion force, $F_{ext}$ is the contact force. The reward is always negative as a discount reward to encourage minimal contact force. However, due to the nature of the task, contact with the environment is unavoidable, so an $S$ shape function is proposed to allow small contact forces while strongly discouraging large ones. A visualization of this reward component is shown in \Cref{fig:reward_f}
    The position, velocity and contact force are normalized by the maximum value allowed for each one. Finally,  $\rho$ is defined as follows:
    \begin{equation}
    \rho = \left\{\begin{matrix}
     500,     & \textrm{Task completed}\\ 
     -200,     & \textrm{Collision} \\
     -1,     & \textrm{Otherwise}
    \end{matrix}\right.
    \label{eq:safety-reward}
    \end{equation}
    The task was considered completed if the Euclidean distance between the robot's end-effector and goal positions was less than 1 mm. The agent is encouraged to complete the task as quickly as possible by discounting the reward for every time step taken. Similar to our previous work \cite{beltranhern2020learning}, we imposed a collision constraint where the agent was penalized for colliding with the environment by giving it a large negative reward and immediately ending the episode. A \textit{collision} is defined as exceeding the force limit $F_{max}$. $F_{max}$ is a hyper-parameter that was defined as 50N in simulation or 30N in the real robot. Lastly, each component was weighted via $w$; all $w$s are hyperparameters. The performance of our new reward signal approach versus the reward signal proposed in our previous work \cite{beltranhern2020learning} is shown in \Cref{apx:reward-types}.

\begin{figure}[ht]
     \centering
     \includegraphics[width=0.7\columnwidth]{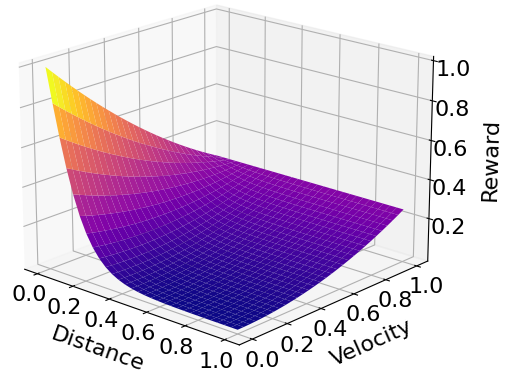}
     \caption{Visualization of the position-velocity-based component of the reward function. $r_{xv}$ in \Cref{eq:reward-position-velocity}}
     \label{fig:reward_xv}
\end{figure}

\begin{figure}[ht]
    \centering
    \includegraphics[width=0.7\columnwidth]{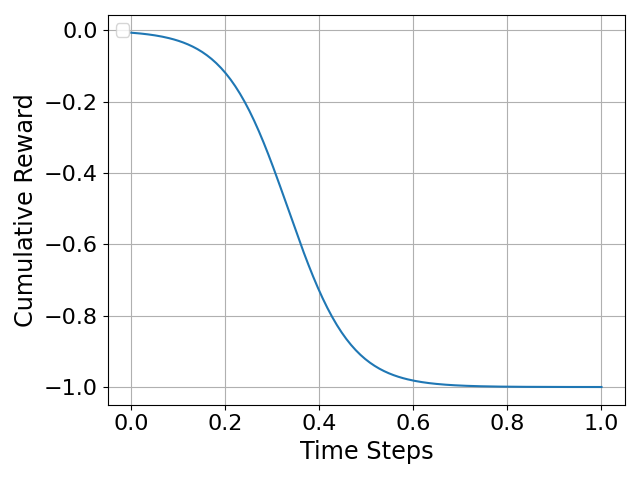}
    \caption{Visualization of the contact-force-based component of the reward function. $r_F$ in \Cref{eq:reward-force}}
    \label{fig:reward_f}
\end{figure}


\subsection{Compliance Control in Task Space}\label{subsec:parallel-control}

    The agent's action space is based on our previous work \cite{beltranhern2020learning}, which consists of learning the force control parameters of a traditional sensor-based force controller. More specifically, we learn the parameters of a parallel position-force controller.
    
    The parallel position-force control requires the fine-tuning of three sets of parameters; gains of a PID for position tracking, gains of a PI for force tracking, and a selection matrix that defines the degree of control between force and position. The controller follows the control law shown in \cref{eq:force-law}

    \begin{equation}\label{eq:force-law}
        \begin{aligned}
            \textbf{x}_c ~=  S (K_p^x\textbf{x}_e &+ K_d^x\Dot{\textbf{x}_e} + \textbf{a}_x)  + (I-S)(K_p^fF_{e}+K_i^f\int F_{e} dt),
        \end{aligned}
    \end{equation}
    where $F_{e} = F_g - F_{ext}$ (goal contact force minus sensed contact force), $\textbf{x}_{e} = \textbf{x}_g - \textbf{x}$ (goal pose minus current pose), $\textbf{a}_x$ represents an arbitrary translation/rotation given by the agent, and $\textbf{x}_c$ is the commanded positions to the robot. The selection matrix is \[S = diag(s_1, ..., s_6),\quad s_j \in [0,1]\]
    The parameters to be learned are $K_p^x$, $K_p^f$, $S$, and $\textbf{a}_x$. One for each of the 6 Cartesian degrees of freedom. The remaining parameters $K_d^x$ and $K_i^f$ were defined proportionally to the $K_p^x$ and $K_p^f$ respectively. Therefore, the action space consists of 24 parameters. Each parameter is bounded to a continuous range of valid values. More details are provided in \cite{beltran2020variable}.
    
    \subsubsection{PID Gains Scheduling} \label{subsubsec:pid-gains-scheduling}
        An additional concept is explored in this paper, PID gains scheduling \cite{vesely2013gain}. Parallel force control for sub-millimeter tolerance insertion tasks tends to get stuck in the region very close to the alignment of the peg onto the hole. In the presence of very small position errors, the PID position controller barely generates any signal. On the other hand, the PI controller overcome the position PID controller due to small contact forces or noise coming from the F/T sensor. Therefore, on insertion tasks with sub-millimeter tolerances, the force controller does not move towards the target pose due to small resistance from the contact with the environment. To address this issue, we introduce PID gains scheduling, where once $\textbf{x}_e$ has been reduced to a position error of less than 1~cm, the PID gains are then scaled up based on the position error $K_p^x=K_p^x*(1/\textbf{x}_e)$. As $\textbf{x}_e$ approaches zero, the value of $K_p^x$ has a hyperbolic growth, thus the value of $K_p^x$ is bounded to be maximum $kn=10$ times its \textit{current} value (the agent's chosen value). \Cref{apx:pid-gains-scheduling} provides a comparison between the use of a traditional PID versus the PID gains scheduling on our proposed method.

    \subsection{Domain Randomization}\label{subsubsec:domain-randomization}

    Domain randomization (DR) \cite{tobin2017domain} is a popular method in robot learning to increase the generalization capabilities of policies trained in simulation, facilitating the transfer of the policy to a real-world robotic agent with minimal to no further refinement of the policy. In principle, the goal of DR is to provide enough variability to the simulated environment during training to generalize better to real-world conditions.
    In robot learning, DR randomizes a set of numerical parameters, $N_r$, of a physics simulator. With each parameter $\psi_i$ being sample from a randomization space $\Psi \in \mathbb{R}^{N_r}$. Each parameter is bounded on a close interval $\{[\psi_i^{low},\psi_i^{high}]\}_{i=1}^{N_r}$. For every episode, a new set of parameters is sample from the randomization space $\psi_i \in \Psi$. The most common approach is to draw sample uniformly from the randomization space.
    In this work, the randomized aspects of the peg-in-hole tasks are defined in \Cref{tab:randomize-values}.
    
    \begin{table*}[h!]
        \centering
        \begin{tabular}{|c|c|c|}
        \hline
        \multicolumn{2}{|c|}{\textbf{Condition}} & \textbf{Set} \\ \hline
        \multirow{2}{*}{\begin{tabular}[c]{@{}c@{}}Initial position \\ (relative to goal)\end{tabular}} & Position (mm) & {[}-50, 50{]} \\ \cline{2-3} 
         & Orientation (\degree) & {[}-30, 30{]} \\ \hline
        \multicolumn{2}{|c|}{\begin{tabular}[c]{@{}c@{}}Peg shape\end{tabular}} & {[}Cylinder, Cuboid, Hexagon prism, Triangular prism{]} \\ \hline
        \multicolumn{2}{|c|}{\begin{tabular}[c]{@{}c@{}}Hole Clearance (mm)\end{tabular}} & {[}\num{3.0}, \num{0.5}{]} \\ \hline
        \multicolumn{2}{|c|}{\begin{tabular}[c]{@{}c@{}} $\epsilon$: Distance from full insertion (mm)\end{tabular}} & {[}\num{15}, \num{1}{]} \\ \hline
        \multicolumn{2}{|c|}{\begin{tabular}[c]{@{}c@{}}Friction\\ (in Gazebo: \textit{surface/friction/ode/mu})\end{tabular}} & {[}\num{1}, \num{5}{]} \\ \hline
        \multicolumn{2}{|c|}{\begin{tabular}[c]{@{}c@{}}Stiffness\\ (in Gazebo: \textit{surface/friction/ode/kp})\end{tabular}} & {[}\num{5.0e-4}, \num{1.0e-6}{]} \\ \hline
        \end{tabular}
        \caption{Domain Randomization parameters and their maximum range of values}
        \label{tab:randomize-values}
    \end{table*}

    \subsection{Curriculum Learning}\label{subsubsec:curriculum-learning}
    
    Curriculum Learning comes from the notion that the order in which information is organized and presented to a learner impacts the learner's performance and training speed. This idea can be observed in the way humans learn, starting with simple concepts and gradually progressing to more complicated problems\cite{peterson2004day,krueger2009flexible}. CL can also be observed in the way we train animals \cite{skinner1958reinforcement}. 
    
    In this work, we follow the notion that starting with easier tasks can help the agent learn better when presented with more difficult tasks later on. We consider the CL problem in the context of DR, where the goal is to reduce the training time by guiding the learning process without loosing domain transferability. Then, the problem becomes how to select parameters from the randomized space $\Psi$ to guide the agent's training. 
    To this end, we consider four main approaches:
    
    \begin{itemize}
        \item Curriculum-based DR: The DR parameter's range of values is determined by the curriculum.
        \item The curriculum's evolution: a linear approach vs an adaptive approach.
        \item The DR sampling strategy: a Uniform distribution (UDR) vs a Gaussian distribution (GDR).
        \item A dynamic reward function based on the curriculum vs a standard reward function.
    \end{itemize}
    
    \subsubsection{Curriculum-based Domain Randomization}\label{subsubsec:dr-cl}
        We tackle the problem of defining a strategy to reduce the complexity of choosing a value for each randomization parameter. Though each parameter of the randomized space $\psi$ can be considered a degree of freedom that can be controlled to define the training tasks, adding new parameters would increases the difficulty of choosing the sequence of tasks to train the agent. Therefore, in order to simplify the problem while preserving the benefits of domain randomization, we propose the following approach: we represent the difficulty level $\mathbf{L}$ of a task as a numerical value in a close interval $[0, 1]$, from easiest to hardest. Then, a sub-set of each randomization parameter $\psi_i$ is defined based on the difficulty level $\mathbf{L_{ep}}$ at the beginning of each episode during training: 
        
        \begin{equation}\label{eq:randomization-ranges}
            \psi_i: [\psi_i^{low}, \psi_i^{low} + \psi_i^{high}*\mathbf{L_{ep}}]
        \end{equation}
        
        where we assume that the parameter's set $\psi_i$ is defined in ascending order, such that, at $low$ and at $high$, the task is relatively the easiest and the hardest, respectively. The parameters considered in this work and their corresponding set are shown in \Cref{tab:randomize-values}. 
    
    \subsubsection{Adaptive Curriculum Learning}\label{subsubsec:adaptive-cl}
        We consider two approaches to update the curriculum difficulty; on the one hand,  the naive approach is to monotonically increase the difficulty in a linear way, regardless of the agent's performance, i.e.,
        
        \begin{equation}
            \mathbf{L_{ep}} = ep/ep_{max}
        \end{equation}
        
        with $\mathbf{L_{ep}}$ being a constraint equal to 1 if the current episode number exceeds a defined maximum number of episodes. On the other hand, we propose an adaptive curriculum based on the agent's performance $\mathbf{P}$ during the last few episodes. 
        The agent's performance is computed as the success rate of the last few episodes. Based on the agent's performance, the curriculum's level is updated by a defined step size $L_{step}$. Two thresholds are also defined. If the agent's performances surpass $L_{thld\_up}$ or fall below $L_{thld\_down}$ such thresholds, then the curriculum's level is increased or decreased respectively. \Cref{alg:cldr} describe our adaptive curriculum approach.

\begin{algorithm}
\caption{Adaptive Curriculum Learning Evolution}\label{alg:cldr}
\begin{algorithmic}
\State $\mathbf{P} = 0$
\For{Every episode $ep$}
    \State Update $\psi$ based on $\mathbf{L_{ep}}$
    \Comment{\cref{eq:randomization-ranges}}
    \State Sample task from $\psi$
    \State $\mathbin{/}\mathbin{/}$ \textit{+1: success, -1: failure}
    \State $\mathbf{P} \mathrel{{+}{=}} $ rollout current policy $\pi$ on task
    \State Update policy
    \If{$\mathbf{P} \ge L_{thld\_up}$}
        \State $L_{ep} \mathrel{{+}{=}}  L_{step}$
        \State $\mathbf{P} = 0$ \Comment{Consider newest rollouts}
    \ElsIf{$\mathbf{P} \le L_{thld\_down}$}
        \State $L_{ep} \mathrel{{-}{=}}  L_{step}$
        \State $\mathbf{P} = 0$ \Comment{Consider newest rollouts}
        \EndIf
\EndFor
\end{algorithmic}
\end{algorithm}

    \subsubsection{Domain Randomization Sampling Strategy}\label{subsubsec:dr-sampling-strategy}
        We consider the type of distribution from which the randomized parameters are sampled. Instead of the typical uniform distribution (UDR), we propose the use of a normal distribution (NDR), $\mathcal{N}(\mu,\sigma^2)$, with the mean being centered around the current curriculum's level $\mathbf{L}_{ep}$, and variance is a hyperparameter. The reason behind this choice is to keep increasing the general difficulty of the task with the increment of the difficulty level, but with a small probability, the curriculum can generate an easier task than the difficulty level to reduce the catastrophic forgetting problem \cite{mccloskey1989catastrophic, french1999catastrophic}. 
        
    \subsubsection{Dynamic Reward Function}\label{subsubsec:dynamic-reward}
        Lastly, for our target task domain, we consider desirable for the RL agent to learn to handle the \textit{hardest} conditions to improve transferability to the real-world environment. To this end, we propose and evaluate a dynamically evolving reward with respect to the curriculum level difficulty. More specifically, the reward $r$, as defined in \Cref{subsubsec:cost-function}, is scaled by the current difficulty level $L_{ep}$; thus, the full reward would be obtained only when the agent reaches and maintains the hardest level.  
    
    \begin{equation}
        \mathbf{r_t^d} = r*\mathbf{L_{ep}}
    \end{equation}
        where $r_t^d$ stands for the dynamic reward at time $t$.
        In other words, at each time step, the reward obtained by the agent is a fraction of the full possible reward for reaching such state. 
\color{black}
\section{Experiments and results} \label{sec:experiments-results}

Through the following experiments, we aimed to understand the performance of our proposed method compared to alternative approaches, in terms of sample efficiency and generalization. To that end, experiments were performed with novel tasks not seen during training in simulation and in the real-world environment, using insertion tasks with medium grade industrial-level tolerances ($\pm 0.01~mm$).

The \textit{baseline} used throughout these experiments was based on our previous work \cite{beltran2020variable}, which mainly focused on the use of DR to enhance domain transferability. This experimental section focus on comparing the different curricula designs, sampling strategies for DR, and curriculum-based dynamic reward. As such, our previous work \cite{beltran2020variable} has been updated to include the new reward function and PID gain scheduling approach, proposed and described in \Cref{subsubsec:cost-function} and \ref{subsubsec:pid-gains-scheduling} respectively, which is used as the \textit{baseline} in this study. Ablation studies of these components of our proposed method are discussed in the Appendix.

\subsection{Experimental Setup} \label{subsec:experimental-setup}
    A simulated environment was used both for training and validation. The Gazebo simulator \cite{koenig2004design} version 9 was used. 
    The choice of simulation environment is discussed in \Cref{sec:discussion}. 
    Two real-world environments were used for validation purpose only; no further re-training was performed on the target domains\footnote{Despite Gazebo’s simulation of the high-stiffness robot being accurate, the robot controllers respond faster than the real robot (maybe due to safety speed reduction on the side of the real robot controller, which we have not modify). Therefore, a minimal calibration is required. From our experience, scaling the reference trajectory or the command send to the controller by a factor of two worked well enough. A rough similarity between the simulation and real robot controller is enough to enable straightforward sim2real transfer.}. The components of the real-world setup is described in \Cref{fig:real-system}. Both environments consist of a Universal Robot 3 e-series robot arm with a control frequency up to 500 Hz. The robotic arm had a force/torque sensor mounted at its end-effector. 
    In the simulation, the peg was considered as part of the robot's end-effector, as shown in \Cref{fig:sim_tasks}. The real-world robot simply used a Robotiq Hand-e parallel gripper. For the toy environments described in \Cref{subsubsec:toy-experiments}, this parallel gripper and a cuboid holder facilitate achieving a strong and stable grasp, similar to the simulation environment. However, for the industrial insertion tasks, we avoided the used of custom-made holders for the real-wold tasks, which increased the difficulty of the tasks as discussed in \Cref{subsubsec:wrs-experiments}.
    
    Our implementation of the RL agent that controlled both the simulated and real robot was developed on top of the Robot Operating System (ROS) \cite{quigley2009ros} with the Universal Robot ROS Driver\footnote{ROS driver for Universal Robot robotic arms developed in collaboration between Universal Robots and the FZI Research Center for Information Technology https://github.com/UniversalRobots/Universal\_Robots\_ROS\_Driver}. In both environments, training of the RL agent was performed on a computer with an Intel i9-10900X CPU and NVIDIA® Quadro RTX™ 8000 GPU. 
    See the accompanying video \footnote{Graphical abstract and experimental results:
    \href{https://youtu.be/_FVQC5OcGjs}{https://youtu.be/\_FVQC5OcGjs}}
    
\begin{figure}[t]
    \centering
    \includegraphics[width=\columnwidth]{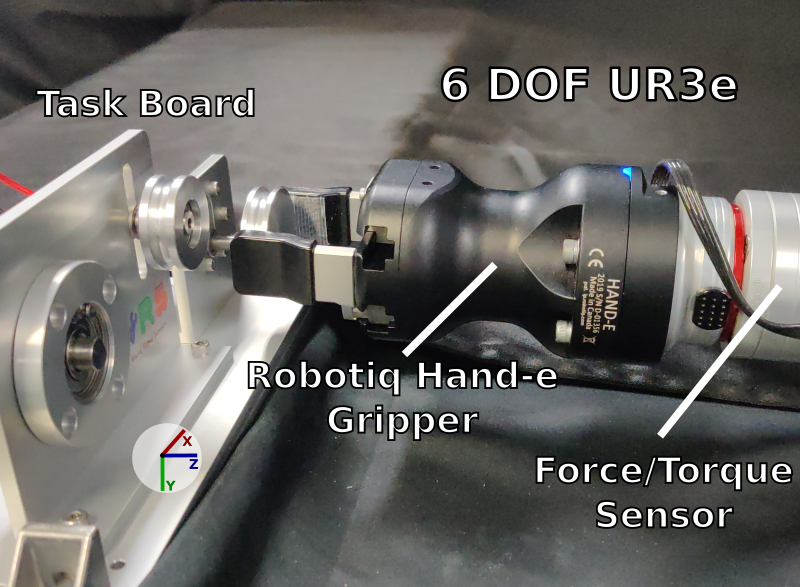}
    \caption{Real experiment environment with a 6-degree-of-freedom UR3e robotic arm. WRS2020 Task board is shown, along side the three insertion tasks used for validation, motor pulley, bearing, and shaft. Each task has industrial level sub-mm tolerances.}
    \label{fig:real-system}
\end{figure}

\subsection{Training}

    The training phase consisted of the repeated execution of the insertion task using a variety of peg shapes and physical parameters of the simulator, as described in \Cref{tab:randomize-values}. An \textit{episode} was defined as a maximum of 1000 time steps, with each step being 50~ms. Early termination of an episode occurs under three conditions; 1) the target goal is reached, the peg inserted, and within $\epsilon$ distance from the full insertion, as described in \Cref{tab:randomize-values}. 2) the robot collides with the task board, i.e., a large contact force is sensed at any point during the task (more than 50~N in simulation or more than 30~N with the real-world robot). 3) the agent gets stuck, thus, the cumulative reward decreases to less than a set value $R_{min}$. 

\begin{figure*}[t]
     \centering
     \hfill
     \begin{subfigure}[b]{0.16\textwidth}
         \centering
         \includegraphics[height=30mm]{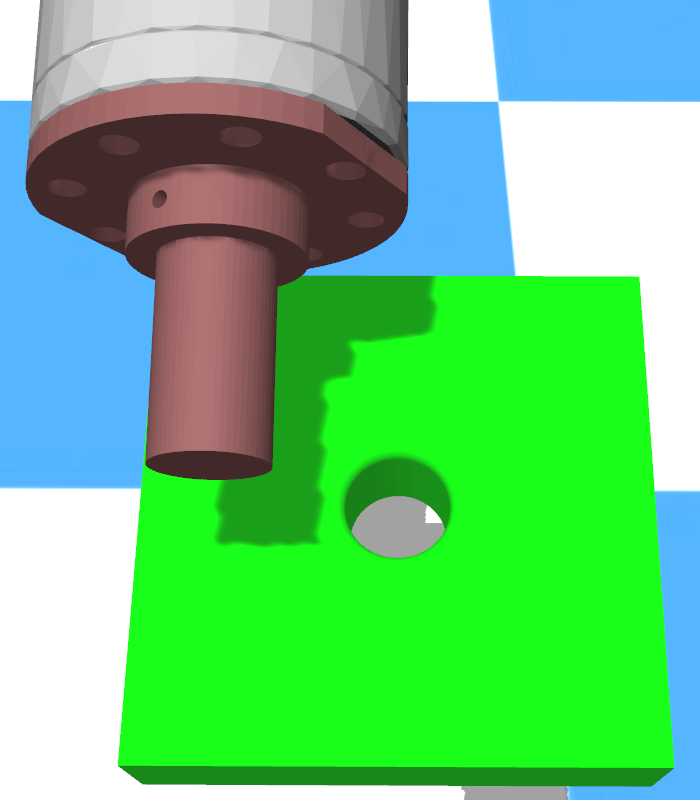}
         \caption{Cylinder.}
         \label{fig:cylinder}
     \end{subfigure}
     \hfill
      \begin{subfigure}[b]{0.16\textwidth}
         \centering
         \includegraphics[height=30mm]{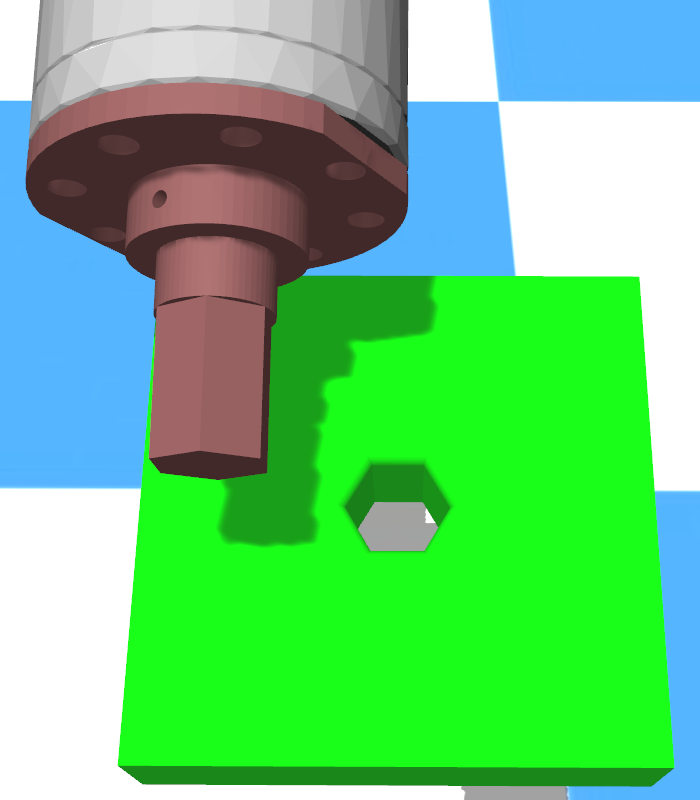}
         \caption{Hexagonal}
         \label{fig:hexagon}
     \end{subfigure}
     \hfill
      \begin{subfigure}[b]{0.16\textwidth}
         \centering
         \includegraphics[height=30mm]{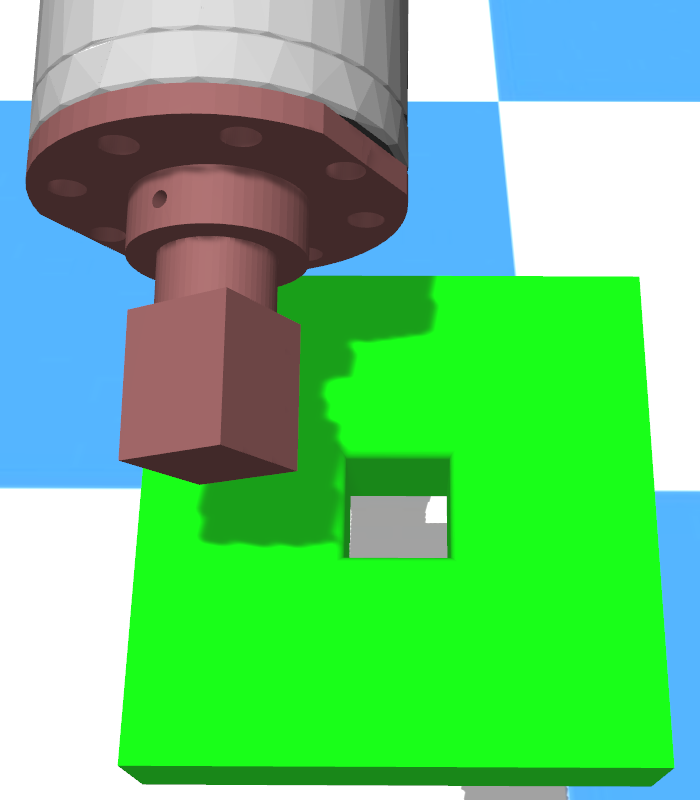}
         \caption{Cuboid}
         \label{fig:cuboid}
     \end{subfigure}
     \hfill
      \begin{subfigure}[b]{0.16\textwidth}
         \centering
         \includegraphics[height=30mm]{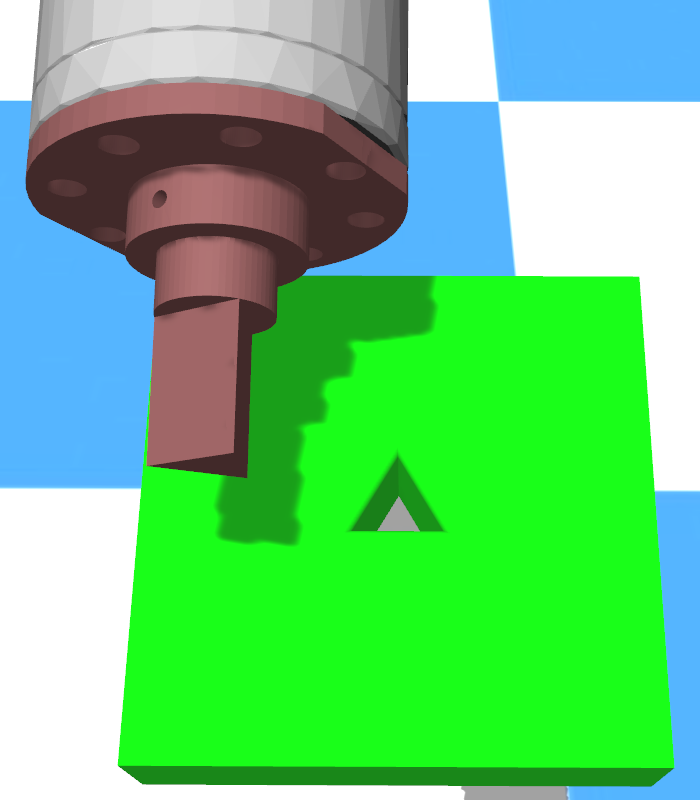}
         \caption{Triangular}
         \label{fig:triangular}
     \end{subfigure}
     \hfill
     \begin{subfigure}[b]{0.16\textwidth}
         \centering
         \includegraphics[height=30mm]{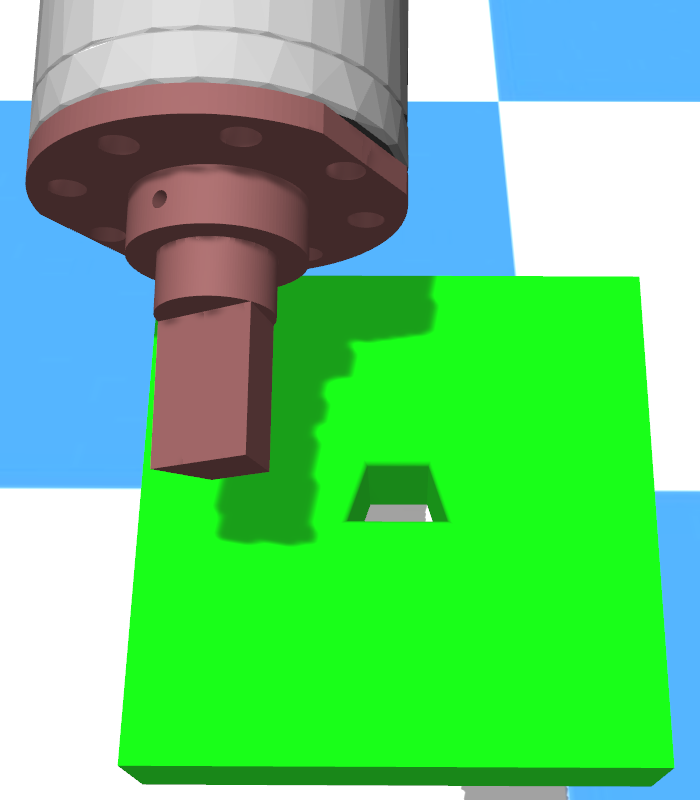}
         \caption{Trapezoid}
         \label{fig:trapezoid}
     \end{subfigure}
     \hfill
     \begin{subfigure}[b]{0.16\textwidth}
         \centering
         \includegraphics[height=30mm]{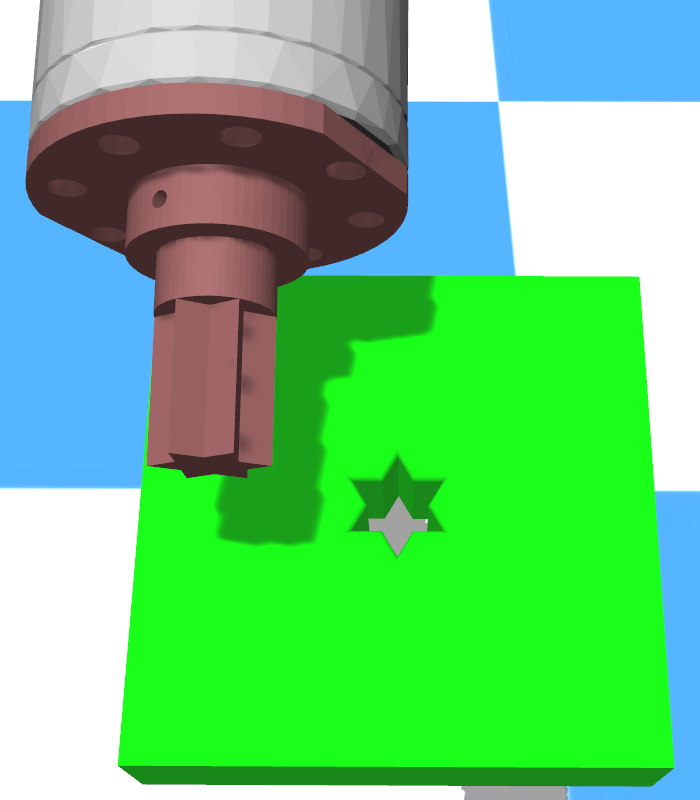}
         \caption{Star}
         \label{fig:star}
     \end{subfigure}
     \hfill
        \caption{Simulated peg-in-hole environments. The cylinder, hexagonal prism, cuboid and triangular prism were used during training. The trapezoid prism and the star prism were used for testing.}
        \label{fig:sim_tasks}
\end{figure*}    

\subsection{Learning performance}

    First, we compare the learning performance of the approaches presented in \Cref{subsubsec:dr-cl}, \ref{subsubsec:adaptive-cl}, and \ref{subsubsec:dr-sampling-strategy}:
    \begin{itemize}
        \item \textit{Baseline}: DR without curriculum learning (No Curriculum), as described in \Cref{sec:experiments-results}.
        \item Linear curriculum with Uniform distribution for DR (Linear Curriculum UDR).
        \item Linear curriculum with Gaussian distribution for DR (Linear Curriculum GDR).
        \item Adaptive curriculum with Uniform distribution for DR (Adp. Curriculum UDR).
        \item Adaptive curriculum with Gaussian distribution for DR (Adp. Curriculum GDR).
    \end{itemize}
    Each training session had a maximum of $100,000$ time steps, one-fifth of the training time required in our previous work \cite{beltran2020variable}.
    As described in \Cref{subsubsec:domain-randomization}, each episode is generated with a different set of values for the randomization parameters.
    \Cref{fig:Learning-comparison} shows the cumulative reward per method during a complete training session. Each training session was repeated with different random seeds. The average value and standard deviation are shown as the bold line and shadow region, respectively. The results are a preliminary highlight of the significant improvement of applying Curriculum Learning compared to the \textit{baseline}, which relied primarily on Domain Randomization alone. Furthermore, the adaptive curricula had a considerable performance above a simple linear increment of the curricula difficulty. Finally, using a Gaussian distribution instead of a Uniform distribution for the sampling of Domain Randomization parameters also significantly improves the agents' performance during learning. The dynamic reward (DyRe) approach discussed in \Cref{subsubsec:dynamic-reward} is not included here as the scale of the reward is different.

\begin{figure}[h!]
    \centering
    \includegraphics[width=\columnwidth]{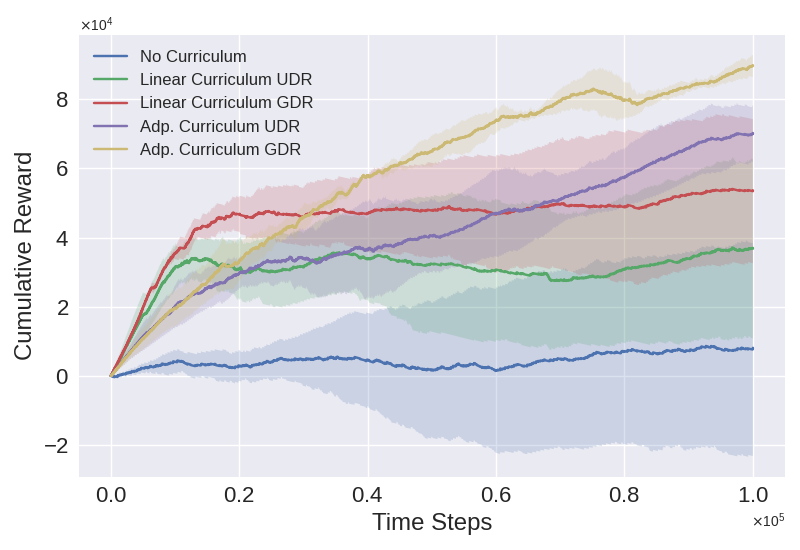}
    \caption{Learning curve comparison using the cumulative reward of the overall training session. Each method was trained three times. The results are aggregated as the average cumulative reward and corresponding standard deviation, represented by the bold line and the shadow region.}
    \label{fig:Learning-comparison}
\end{figure}

\subsection{Evaluating learned Policies} \label{sucsec:eval-learning-curves}

    Next, we evaluated the performance of the learned policies on novel conditions not seen during training. Each policy was executed 100 times with different initial conditions and randomized parameters (with a fixed random seed for a fair comparison). More specifically, the peg shapes used for testing were a trapezoid prism and star prism, as shown in \Cref{fig:sim_tasks}. The trapezoid introduces a non-symmetric-shaped peg. The star prism peg is more challenging due to its sharp corners that make the peg prone to getting stuck during the aligning phase, making the overall insertion task harder to complete during the allowed time limit of 50 seconds (same time limit as during training).
    
    The results are shown in \Cref{tab:sim-results}. They include a comparison of the overall success rate and the average time needed to complete the task; failure cases are not included in the computation of the average time. Two main conclusions can be drawn from these results; 1) A curriculum may seem to have a better learning performance, but the resulting policy may not transfer well to novel environments, as is the case with the \textit{Linear Curricula} methods. Such linear curriculum approaches performed just slightly better than not using a curriculum at all. 2) The most successful methods are not necessarily the fastest. Our simulation environment did not handle very well friction between the peg and the task board, due to the high stiffness of the robot joints. In this almost friction-less world, the \textit{Adp. Curriculum UDR} method is able to solve the tasks between 20\% to 50\% faster than our best method \textit{Adp. Curriculum GDR DyRe}. However, the success rate of our method is at least 19\% higher.
    The main reason for such results is that since contact force and collision avoidance have higher priority than speed during learning, our proposed method moves slower when the peg gets closer to the task board so the contact force is reduced. This conclusion is further supported by the real-world experimental data described next in \Cref{subsubsec:learning-force-control}.

\begin{table}[h!]
\centering
\begin{tabular}{|c|cc|cc|}
\hline
\multirow{2}{*}{\textbf{Method}} & \multicolumn{2}{c|}{\textbf{Trapezoid   Prism}} & \multicolumn{2}{c|}{\textbf{Star Prism}} \\ \cline{2-5} 
 & \multicolumn{1}{c|}{\begin{tabular}[c]{@{}c@{}}Success\\ Rate\end{tabular}} & \begin{tabular}[c]{@{}c@{}}Avg.\\ Time(s)\end{tabular} & \multicolumn{1}{c|}{\begin{tabular}[c]{@{}c@{}}Success\\ Rate\end{tabular}} & \begin{tabular}[c]{@{}c@{}}Avg.\\ Time(s)\end{tabular} \\ \hline
No Curriculum & \multicolumn{1}{c|}{0.88} & 9.585 & \multicolumn{1}{c|}{0.627} & 11.304 \\ \hline
Linear Curriculum UDR & \multicolumn{1}{c|}{1.00} & 9.817 & \multicolumn{1}{c|}{0.696} & 9.152 \\ \hline
Linera Curriculum GDR & \multicolumn{1}{c|}{1.00} & 11.650 & \multicolumn{1}{c|}{0.775} & 11.780 \\ \hline
Adp. Curriculum UDR & \multicolumn{1}{c|}{1.00} & \textbf{6.881} & \multicolumn{1}{c|}{0.706} & \textbf{6.875} \\ \hline
Adp. Curriculum GDR & \multicolumn{1}{c|}{1.00} & 8.460 & \multicolumn{1}{c|}{0.794} & 8.013 \\ \hline
\begin{tabular}[c]{@{}c@{}}Adp. Curriculum \\ UDR DyRe\end{tabular} & \multicolumn{1}{c|}{1.00} & 8.429 & \multicolumn{1}{c|}{0.873} & 11.602 \\ \hline
\textbf{\begin{tabular}[c]{@{}c@{}}Adp. Curriculum \\ GDR DyRe\end{tabular}} & \multicolumn{1}{c|}{1.00} & 8.400 & \multicolumn{1}{c|}{\textbf{0.902}} & 11.544 \\ \hline
\end{tabular}
\caption{Evaluation of learned policies on novel conditions.}
\label{tab:sim-results}
\end{table}

\subsection{Real-world experiments} \label{subsec:real-world-experiments}

\begin{figure*}[t]
     \centering
     \begin{subfigure}[b]{0.39\textwidth}
         \centering
         \includegraphics[width=\columnwidth]{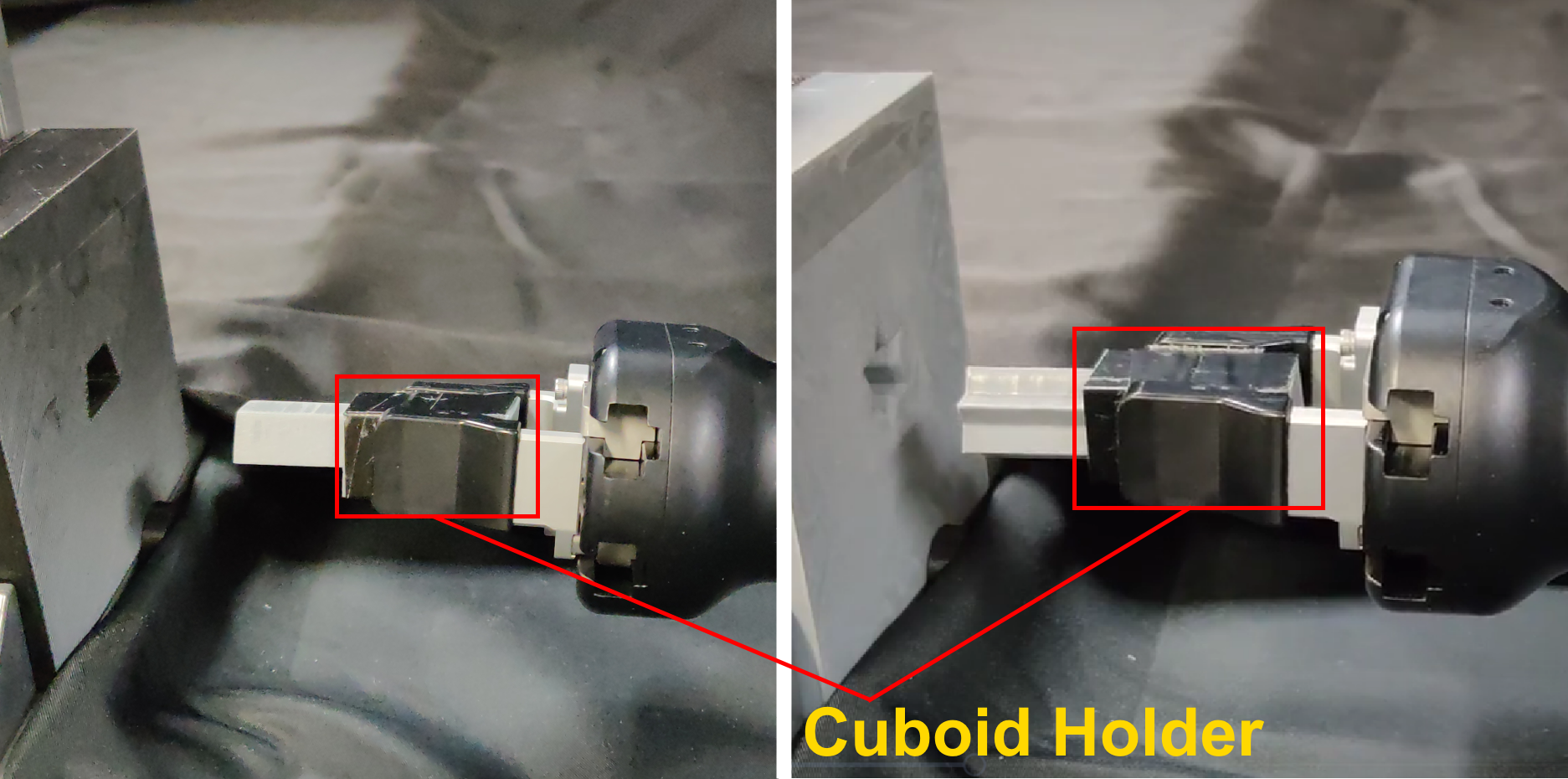}
         \caption{Trapezoid and star prism pegs}
         \label{fig:toy_tasks}
     \end{subfigure}
     \hfill
     \begin{subfigure}[b]{0.59\textwidth}
         \centering
         \includegraphics[width=\columnwidth]{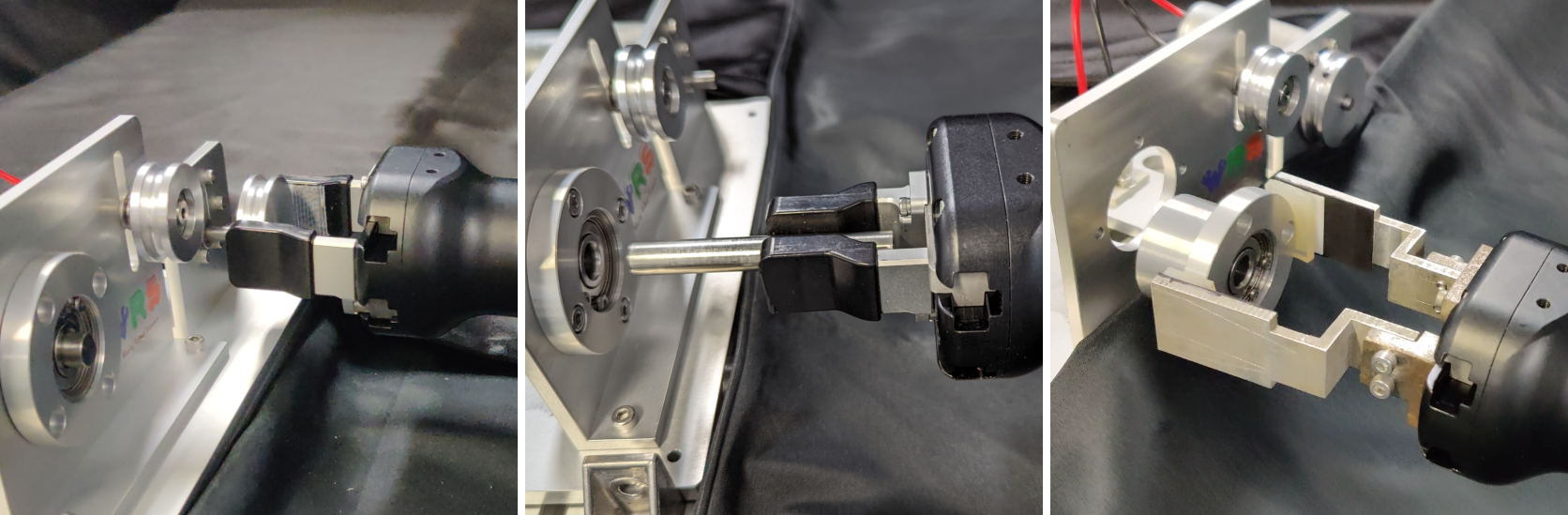}
         \caption{Motor Pulley, Shaft, and Bearing.}
         \label{fig:wrs_tasks}
     \end{subfigure}
     \hfill
        \caption{Real-world experimental scenarios. \textbf{Left}: 3D printed primitive shape-pegs, different from the ones used for training in simulation. \textbf{Right}: Industrial level insertion tasks from the WRS2020 Robotics Assembly Challenge \cite{wrs2020_rulebook}.}
        \label{fig:toy_experiments}
\end{figure*}

    We performed two sets of experiments to evaluate the transferability of the learned policies to the real world and to novel tasks. The experiments were performed using the \textit{baseline} (No Curriculum) method and our newly proposed method (Adp. Curriculum DyRe GDR), which achieved the best results from the evaluation in simulation. 
    The first set of tasks consisted of the same trapezoid and star prism-shaped pegs as the simulation experiments, which were not presented during training. A simplified peg was 3D printed using PLA material, as shown in \Cref{fig:toy_tasks}. 
    The second set of tasks consisted of novel industrial level insertion tasks (See \Cref{fig:wrs_tasks}), similarly, these tasks were unseen during training in simulation. Both sets of tasks had sub-millimeter tolerances. Thirty trials were performed per method and task. The success rate and average completion time were measured where the initial position and orientation of the robot's end-effector at each trial were different and randomly sampled from a fixed random seed to fairly compare both methods. Each trial had a 500-time steps limit, i.e., 25~s.  
    
\subsubsection{Primitive Shaped Pegs} \label{subsubsec:toy-experiments}

    The 3D printed pegs were designed with a cuboid holder, as shown in \Cref{fig:toy_tasks}, to increase the stability of the grasp. As a result, the stiffness of the contact is very high, as the task board was also firmly fixed to the workspace. Additionally, there is high friction due to the PLA material used for 3D printing and the imperfections on the printed surface. The high stiffness and friction made the task challenging. The results are shown in \Cref{tab:real-toy-results}. As mentioned before, for this test, the \textit{baseline} was also trained with only one-fifth of the samples shown to be needed \cite{beltran2020variable} to learn a successful policy. Thus, the learned policy's less refined force control tends to apply too much force to complete the task quickly, but the high friction and the corners of the star-shaped hole cause the peg to get stuck easily. Therefore, the \textit{baseline} method struggled to align the star prism peg and to get unstuck. On the other hand, our newly proposed approach successfully adapted to the real-world environment and succeeded at the novel tasks without further re-training the policies, just a straightforward sim-to-real transfer.

\begin{table}[h!]
\centering
\begin{tabular}{|c|cc|cc|}
\hline
\multirow{2}{*}{Method} & \multicolumn{2}{c|}{Trapezoid Prism Peg} & \multicolumn{2}{c|}{Star Prism Peg} \\ \cline{2-5} 
 & \multicolumn{1}{c|}{\begin{tabular}[c]{@{}c@{}}Success\\ Rate\end{tabular}} & \begin{tabular}[c]{@{}c@{}}Avg. \\ Time(s)\end{tabular} & \multicolumn{1}{c|}{\begin{tabular}[c]{@{}c@{}}Success\\ Rate\end{tabular}} & \begin{tabular}[c]{@{}c@{}}Avg. \\ Time(s)\end{tabular} \\ \hline
No Curriculum & \multicolumn{1}{c|}{1.000} & 6.500 & \multicolumn{1}{c|}{0.000} & - \\ \hline
\begin{tabular}[c]{@{}c@{}}Adp. Curriculum \\ DyRe GDR\end{tabular} & \multicolumn{1}{c|}{\textbf{1.000}} & \textbf{5.465} & \multicolumn{1}{c|}{\textbf{1.000}} & \textbf{6.023} \\ \hline
\end{tabular}
\caption{Evaluating learned policies on the real-world environment, using 2 toy scenarios not seen during training on simulation.}
\label{tab:real-toy-results}
\end{table}


\subsubsection{Industrial Level Insertion Tasks} \label{subsubsec:wrs-experiments}

    The second set of tasks used for evaluation consisted of 3 insertion tasks with industrial level tolerances. These were chosen from the assembly task used in the Industrial Robots Assembly Challenge of the World Robot Summit 2020 edition \cite{wrs2020_rulebook}. The tasks, as shown in \Cref{fig:wrs_tasks}, were the insertion of a pulley into a motor shaft, a shaft into a bearing, and a bearing into a plate. Similar to the previous tasks, the learned policies were directly transferred from the simulation environment without further training.
    These tasks are considerably more challenging as the grasp's stability significantly impacts the success. All three manipulated objects are round and grasped directly with a standard parallel gripper. Thus, torques applied along the direction of the grasp could easily change the object's orientation in the gripper. Small orientation changes significantly affect these very tight insertion tasks.
    
    The results are shown in \Cref{tab:real-wrs-results}. 
    Our newly proposed method (Adp. Curriculum DyRe GDR) achieved a high success rate in all the tasks. For the motor pulley and the shaft tasks, our method also solves the task faster by finding the right fit faster. Our method is less likely to get stuck as it applies less contact force as shown in \Cref{fig:distance-force-comparison} and \ref{fig:distance-force-comparison-failure}
    In the case of the bearing task, the \textit{baseline} method, when successful, is slightly faster as it tends to apply higher contact force and move faster once the parts are aligned. However, the same high contact force makes it harder to find the proper alignment, thus resulting in a very low success rate. As a result, our newly proposed method outperforms the \textit{baseline} method, achieving a much higher success rate. These results are better appreciated in the supplemented video\footnote{Supplemental video:
    \href{https://youtu.be/_FVQC5OcGjs}{https://youtu.be/\_FVQC5OcGjs}}.
    
    %
    

\begin{table*}[h!]
\centering
\begin{tabular}{|c|cc|cc|cc|}
\hline
\multirow{2}{*}{Method} & \multicolumn{2}{c|}{Motor   Pulley} & \multicolumn{2}{c|}{Shaft} & \multicolumn{2}{c|}{Bearing} \\ \cline{2-7} 
 & \multicolumn{1}{c|}{\begin{tabular}[c]{@{}c@{}}Success\\ Rate\end{tabular}} & \begin{tabular}[c]{@{}c@{}}Avg.\\ Time(s)\end{tabular} & \multicolumn{1}{c|}{\begin{tabular}[c]{@{}c@{}}Success\\ Rate\end{tabular}} & \begin{tabular}[c]{@{}c@{}}Avg.\\ Time(s)\end{tabular} & \multicolumn{1}{c|}{\begin{tabular}[c]{@{}c@{}}Success\\ Rate\end{tabular}} & \begin{tabular}[c]{@{}c@{}}Avg.\\ Time (s)\end{tabular} \\ \hline
No Curriculum & \multicolumn{1}{c|}{0.400} & 8.258 & \multicolumn{1}{c|}{0.667} & 9.199 & \multicolumn{1}{c|}{0.267} & 6.819 \\ \hline
\begin{tabular}[c]{@{}c@{}}Adp. Curriculum \\ DyRe GDR\end{tabular} & \multicolumn{1}{c|}{0.867} & 7.250 & \multicolumn{1}{c|}{0.833} & 7.015 & \multicolumn{1}{c|}{0.700} & 7.212 \\ \hline
\end{tabular}
\caption{Evaluating learned policies on the real-world environment, using 2 toy scenarios not seen during training on simulation.}
\label{tab:real-wrs-results}
\end{table*}

\subsubsection{Learning Force Control} \label{subsubsec:learning-force-control}

    In addition to the success rate and time to completion, we compare the detailed performance of the two methods. \Cref{fig:distance-force-comparison} shows the performance of both methods side by side for the three industrial insertion tasks. 
    For simplicity, only the z-axis (i.e., the insertion direction), distance error (mm), and contact force (N) are displayed. 
    As shown in \Cref{fig:distance-force-comparison}, our proposed method is more time-efficient and applies less contact force to the coupling part. Less contact force is desirable to avoid damage to either the assembly part or the robot.
    Similarly, \Cref{fig:distance-force-comparison-failure} shows the comparison of trials where both agents fail to complete the task on time. Though both agents failed, our method again shows a reduced exertion of contact force. 
    In both cases, our method applies about 30\% less contact force.
    
\begin{figure*}[h!]
     \centering
     \begin{subfigure}[b]{0.32\textwidth}
         \centering
         \includegraphics[width=\columnwidth]{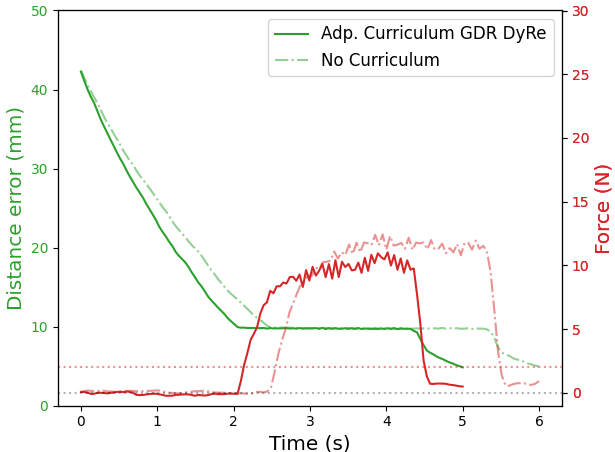}
         \caption{Motor Pulley}
         \label{fig:motor_pulley_f}
     \end{subfigure}
     \hfill
     \begin{subfigure}[b]{0.32\textwidth}
         \centering
         \includegraphics[width=\columnwidth]{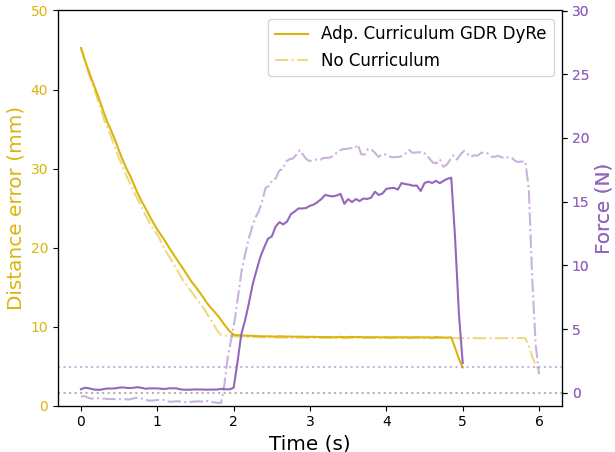}
         \caption{Shaft}
         \label{fig:shaft_f}
     \end{subfigure}
     \hfill
     \begin{subfigure}[b]{0.32\textwidth}
         \centering
         \includegraphics[width=\columnwidth]{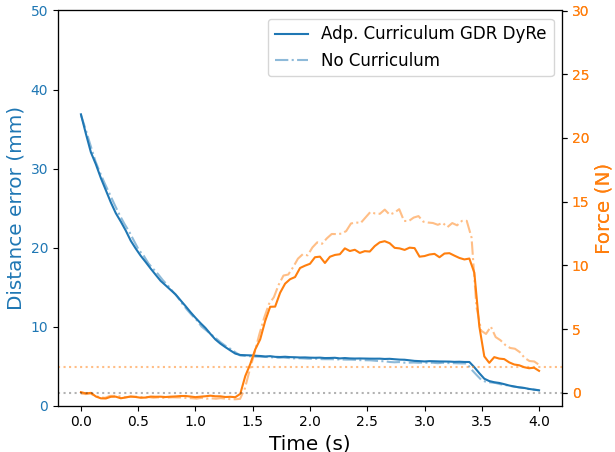}
         \caption{Bearing}
         \label{fig:bearing_f}
     \end{subfigure}
        \caption{Agents performance on WRS2020 insertion tasks. For clarity, only the z-axis (Insertion direction) distance error and contact force are displayed. Comparison was made for each task when both methods successfully completed the task.}
        \label{fig:distance-force-comparison}
\end{figure*}

\begin{figure*}[h!]
     \centering
     \begin{subfigure}[b]{0.32\textwidth}
         \centering
         \includegraphics[width=\columnwidth]{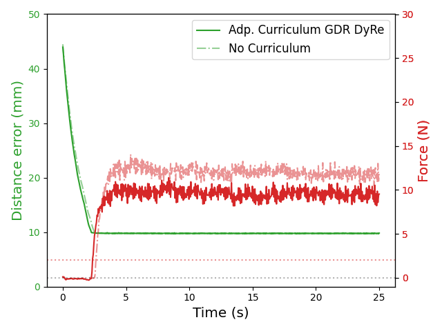}
         \caption{Motor Pulley}
         \label{fig:motor_pulley}
     \end{subfigure}
     \hfill
     \begin{subfigure}[b]{0.32\textwidth}
         \centering
         \includegraphics[width=\columnwidth]{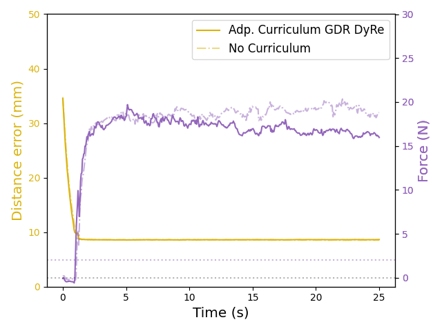}
         \caption{Shaft}
         \label{fig:shaft}
     \end{subfigure}
     \hfill
     \begin{subfigure}[b]{0.32\textwidth}
         \centering
         \includegraphics[width=\columnwidth]{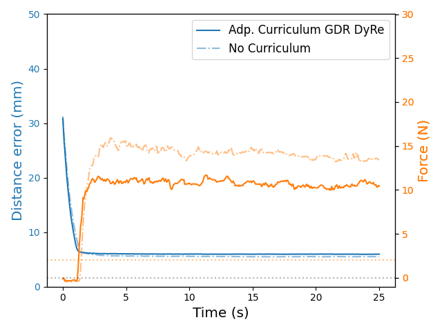}
         \caption{Bearing}
         \label{fig:bearing}
     \end{subfigure}
        \caption{Performance of both methods where both failed to complete the task within the time limit.}
        \label{fig:distance-force-comparison-failure}
\end{figure*}


    



\section{Discussion}
\label{sec:discussion}
    Training a reinforcement learning agent with a curriculum that starts from easier tasks with reduced risk of encountering fatal states (e.g., a collision during a manipulation task) improves the learning sample efficiency and overall performance. In this case study, in particular, we integrate CL to contact-rich peg insertion tasks. A task is defined by various physical parameters as described in \Cref{tab:randomize-values}.
    We aim to allow the agent to carefully explore more states by presenting tasks in increasing order of difficulty, e.g., by reducing the stiffness of the contact between the peg and the board, the agent is less likely to apply excessive contact force to the environment (i.e., a collision).
    At the same time, starting with easier tasks, such as a shorter distance from the initial position to goal one, reduces the overall exploration. The curriculum is design around a domain randomization approach to preserve and enhance the domain transferability benefits from DR. Nevertheless, the type of curricula is very relevant for achieving better performance, both in terms of sample efficiency and success rate, as seen in the results in \cref{subsubsec:toy-experiments}.
    
    From our previous work \cite{beltran2020variable}, we demonstrated that with sufficient domain randomization-based training in simulation (at least $500,000$ time steps or about 8 hours) and further retraining in the real-robot environment, it is possible to learn policies that adapt to novel domains successfully. The present paper introduce a study on Curriculum Learning to tackle the problem of sample-efficiency, and the need to retrain in the target domain. The experimental results on the real-robot environment confirms that our newly proposed method is effective in learning contact-rich force control tasks. Despite that our proposed method was trained only in simulation with one-fifth of the training time used in our previous work \cite{beltran2020variable} and without further retraining, it achieves a high success rate on novel tasks, including challenging industrial insertion tasks with sub-millimeter tolerances.
    
    The main limitation of our proposed method is the assumption that the domain randomization parameter ranges are organized in order from \textit{easy} to \textit{difficult}. Prior knowledge is required to determine the \textit{difficulty} of a physical parameters on a given task. Such prior knowledge is task-specific and not necessarily easy to obtain or even impossible to determine manually. For example, considering only the peg-in-hole task and the stiffness between the peg and the task board, a low-stiffness can intuitively seem easier to handle for a high-stiffness robot arm, as less careful force control is required to achieve the task without generating large contact forces. However, depending on the material and how low the stiffness is, the peg may get stuck, or the task board may be deformed to such an extent that the task becomes impossible to solve.
    To tackle this problem, an interesting future avenue is to add a layer of learning, following works such as \cite{mehta2020adr,svetlik2017automatic}. The main idea of this line of research work is to train a neural network to define the RL agent's tasks. In other words, the network learns to choose the best values for the randomization parameters at each episode to increase the performance of the learned policy. The approaches differ in how the new network is trained, and how the agent's performance is defined, such as the cumulative reward, success rate, or another evaluation metric.
    Following these self-learned curricula approaches reduces the burden on prior knowledge, though as discussed in \cite{mehta2020adr}, a self-learned curriculum can provide an insight into incompatibilities between the task and randomization ranges. Therefore, such approaches may allow the use of many other parameters of the physics simulator for domain randomization, potentially increasing the transferability to novel domains. Nevertheless, the possible downside is the requirement of longer training sessions due to the added complexity. 
    
    Another future avenue to further improve the presented work is the choice of a simulation environment. At the time of writing this paper, there are various physics engine simulators available that simulate the contact dynamics between bodies with different degrees of accuracy, among other capabilities. Our choice of the Gazebo simulator was motivated by its realistic simulation of rigid position-controlled robots. Additionally, the availability of ROS controllers that worked the same in the simulated and the real-world robot reduces the implementation burden and facilitates sim2real transfer. Nonetheless, other simulators provide better contact dynamics and are better adapted for Reinforcement Learning applications, such as Mujoco \cite{todorov2012mujoco}, or Nvidia Isaac Sim \cite{liang2018gpu}. Working with such simulators would be a significant improvement, as Domain Randomization is also easier to implement for vision-based learning methods and physical parameters of the simulated environment. 
    
\section{Conclusions}
    This paper has studied the application of different approaches that combine Curriculum Learning with Domain Randomization to learn contact-rich manipulation tasks, particularly assembly tasks such as peg insertion. Based on such a study, we proposed to improve sample efficiency and generalization by training an agent purely in simulation, with the training being guided with CL and enhanced with DR. Additionally, this work introduced two enhancements to our learning framework, a new dense reward function and a PID gain scheduling approach, described in \Cref{subsubsec:cost-function} and \ref{subsubsec:pid-gains-scheduling} respectively, and validated in the Appendix.

    The learning framework proposed in this work is based on our previous work \cite{beltran2020variable} where a combination of sim2real with DR was proposed. Our previous methods still required a considerably large amount of agent's interaction with its environment and additional refinement in the real-world environment to learn a robust policy. On the contrary, our novel approach can be trained purely in simulation with only toy insertion tasks. 
    Empirical results showed that with our proposed method a successful policy can be learned using only one-fifth of the samples needed in our previous work. Such policies can be straightforwardly transferred to real-world environments and still achieve a high success rate, up to 86\%, on novel complex industrial insertion tasks not seen during training.

\bibliographystyle{ieeetr}
\bibliography{main}





\clearpage

\appendix
\section{Ablation studies} \label{appendix}
    In this section we compare the performance of the proposed method where the improvements presented in this work; the new dense reward function (\cref{subsubsec:cost-function}), and the addition of a PID gains scheduling to the force controller (\cref{subsec:parallel-control}). Similar to the experimental setup described in \Cref{sucsec:eval-learning-curves}, each method was evaluate on simulation by executing their corresponding learned policy over a 100 trials for each task.

\subsection{Reward Functions} \label{apx:reward-types}

     The newly proposed dense reward function includes the robot's end-effector velocity combined with the error position. Our aim is to encourage the agent to move faster while being far from the target pose, but to move slower when closer to the target position to reduce the risk of high contact forces. 
     For a fair comparison, the implementation of the \textit{Old reward function} method and our proposed \textit{New reward function} were identical except for the type of reward function. Both methods are based on our proposed method (Adp. Curriculum GDR DyRe), as described in \Cref{subsubsec:curriculum-learning}.
    \Cref{fig:reward-types} shows the comparison of the training session, and the overall cumulative reward for each method. In addition, both approaches were tested on two novel tasks on simulation, the same tasks described in \Cref{sucsec:eval-learning-curves}. The results, displayed in \Cref{tab:reward-types}, shows a significant improvement in performance. Our approach using the \textit{New reward} achieved a higher success rate. Additionally, our approach also was more efficient at solving the tasks. In average, the task are solved at least twice as fast.

\begin{figure}[h!]
    \centering
    \includegraphics[width=\columnwidth]{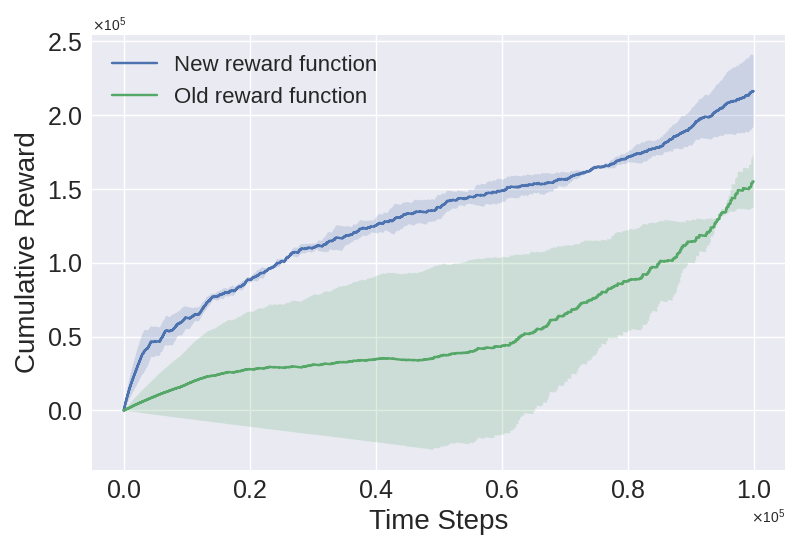}
    \caption{Learning curve comparison.}
    \label{fig:reward-types}
\end{figure}
\begin{table}[h!]
\centering
\begin{tabular}{|c|cc|cc|}
\hline
\multirow{2}{*}{Method} & \multicolumn{2}{c|}{\begin{tabular}[c]{@{}c@{}}Trapezoid \\ Prism Peg\end{tabular}} & \multicolumn{2}{c|}{\begin{tabular}[c]{@{}c@{}}Star\\ Prism Peg\end{tabular}} \\ \cline{2-5} 
 & \multicolumn{1}{c|}{\begin{tabular}[c]{@{}c@{}}Success\\  Rate\end{tabular}} & \begin{tabular}[c]{@{}c@{}}Avg. \\ Time(s)\end{tabular} & \multicolumn{1}{c|}{\begin{tabular}[c]{@{}c@{}}Success \\ Rate\end{tabular}} & \begin{tabular}[c]{@{}c@{}}Avg. \\ Time(s)\end{tabular} \\ \hline
\begin{tabular}[c]{@{}c@{}}Old \\ Reward\end{tabular} & \multicolumn{1}{c|}{0.98} & 14.979 & \multicolumn{1}{c|}{0.85} & 12.271 \\ \hline
\begin{tabular}[c]{@{}c@{}}New\\  Reward\end{tabular} & \multicolumn{1}{c|}{\textbf{1.000}} & \textbf{5.465} & \multicolumn{1}{c|}{\textbf{1.000}} & \textbf{6.023} \\ \hline
\end{tabular}
\caption{Success rate on novel tasks on the simulated environment.}
\label{tab:reward-types}
\end{table}
 
\unskip
\newpage
\subsection{Force Controller Position PID types} \label{apx:pid-gains-scheduling}

     Furthermore, we updated the PID position controller of our force controller to enhance the performance of the agent when the position error is very small. Similarly, both method were identical except for the implementation of the PID position controller and based on our proposed method (Adp. Curriculum GDR DyRe), as described in \Cref{subsubsec:curriculum-learning}. 
     The results of evaluating each method on novel tasks not seen during training are shown in \Cref{fig:pid-gains-scheduling}, for the learning curve, and in \Cref{tab:pid-gains-scheduling}.
     The results show a considerable improvement of performance when using our proposed \textit{PID gain scheduling} approach. The success rate achieved by our \textit{PID gain scheduling} approach is about twice the achieved with the \textit{Normal PID} approach. On top of that, our \textit{PID gain scheduling} approach can solve the tasks in less than half the time required by the \textit{Normal PID} approach.
     
\begin{figure}[h!]
    \centering
    \includegraphics[width=\columnwidth]{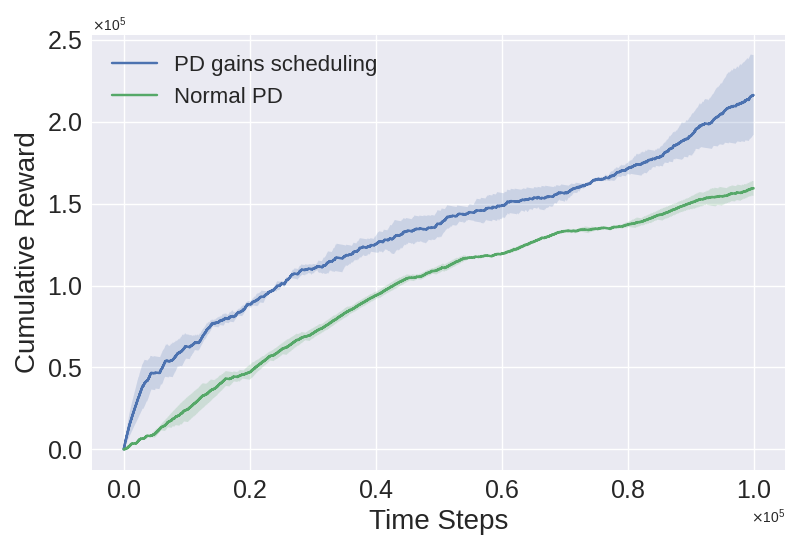}
    \caption{Learning curve comparison.}
    \label{fig:pid-gains-scheduling}
\end{figure}
\begin{table}[h!]
\centering
\begin{tabular}{|c|cc|cc|}
\hline
\multirow{2}{*}{Method} & \multicolumn{2}{c|}{\begin{tabular}[c]{@{}c@{}}Trapezoid \\ Prism Peg\end{tabular}} & \multicolumn{2}{c|}{\begin{tabular}[c]{@{}c@{}}Star\\ Prism Peg\end{tabular}} \\ \cline{2-5} 
 & \multicolumn{1}{c|}{\begin{tabular}[c]{@{}c@{}}Success\\  Rate\end{tabular}} & \begin{tabular}[c]{@{}c@{}}Avg. \\ Time(s)\end{tabular} & \multicolumn{1}{c|}{\begin{tabular}[c]{@{}c@{}}Success \\ Rate\end{tabular}} & \begin{tabular}[c]{@{}c@{}}Avg. \\ Time(s)\end{tabular} \\ \hline
\begin{tabular}[c]{@{}c@{}}Normal \\ PID\end{tabular} & \multicolumn{1}{c|}{0.780} & 12.798 & \multicolumn{1}{c|}{0.500} & 13.949 \\ \hline
\begin{tabular}[c]{@{}c@{}}PID gains\\  Scheduling\end{tabular} & \multicolumn{1}{c|}{\textbf{1.000}} & \textbf{5.465} & \multicolumn{1}{c|}{\textbf{1.000}} & \textbf{6.023} \\ \hline
\end{tabular} 
\caption{Success rate on novel tasks on the simulated environment.}
\label{tab:pid-gains-scheduling}
\end{table}






\end{document}